\def\year{2022}\relax
\documentclass[letterpaper]{article} % DO NOT CHANGE THIS
\usepackage{aaai22}  % DO NOT CHANGE THIS
\usepackage{times}  % DO NOT CHANGE THIS
\usepackage{helvet}  % DO NOT CHANGE THIS
\usepackage{courier}  % DO NOT CHANGE THIS
\usepackage[hyphens]{url}  % DO NOT CHANGE THIS
\usepackage{graphicx} % DO NOT CHANGE THIS
\usepackage{natbib}  % DO NOT CHANGE THIS AND DO NOT ADD ANY OPTIONS TO IT
\usepackage{caption} % DO NOT CHANGE THIS AND DO NOT ADD ANY OPTIONS TO IT
\DeclareCaptionStyle{ruled}{labelfont=normalfont,labelsep=colon,strut=off} % DO NOT CHANGE THIS
\title{Mapping Knowledge Representations to Concepts: \\ A Review and New Perspectives}
\author {
    Lars Holmberg,\textsuperscript{\rm 1}\thanks{This work was partially financed by the Knowledge Foundation through the Internet of Things and People research profile.}
    Paul Davidsson, \textsuperscript{\rm 1}
    Per Linde \textsuperscript{\rm 2}
}
\affiliations {
    \textsuperscript{\rm 1} Department of Computer Science and Media Technology\\
    \textsuperscript{\rm 2} School of Arts and Communication\\ Malmö University, Sweden\\
    lars.holmberg@mau.se, paul.davidsson@mau.se, per.linde@mau.se
}
\usepackage{amsmath,amssymb,amsfonts}
\usepackage[british]{babel}
\usepackage[autostyle]{csquotes}
\usepackage{array}
\usepackage{multirow}
\usepackage{wrapfig}
\begin{document}
\maketitle
\begin{abstract}
The success of neural networks builds to a large extent on their ability to create internal knowledge representations from real-world high-dimensional data, such as images, sound, or text. Approaches to extract and present these representations, in order to explain the neural network's decisions, is an active and multifaceted research field. To gain a deeper understanding of a central aspect of this field, we have performed a targeted review focusing on research that aims to associate internal representations with human understandable concepts. In doing this, we added a perspective on the existing research by using primarily deductive nomological explanations as a proposed taxonomy. We find this taxonomy and theories of causality, useful for understanding what can be expected, and not expected, from neural network explanations. The analysis additionally uncovers an ambiguity in the reviewed literature related to the goal of model explainability; is it understanding the ML model or, is it actionable explanations useful in the deployment domain?
\end{abstract}

\section{Introduction}
Digitalisation influences all parts of society. Even the meaning of the central philosophical term episteme has shifted from being best translated as knowledge, towards being best translated as understanding~\cite{sep-understanding}. The shift can be exemplified by an ever-present weather app that knows if and when it will rain but has no understanding, in a human sense, concerning why it rains or the subjective implications for an individual human being. Focus in this work is then on the tension between the prospective human understander and the third-person objectivising stance characteristic of the natural sciences~\cite{Grimm2016}, here represented by Artificial Intelligence (AI) and in particular Machine Learning (ML). 

Explanations, typically answering a \textit{Why} or \textit{What if} question, is a common human way to bring understanding of the natural world or other people~\cite{sep-understanding}. Bridging the gap, between information processing systems, like AI/ML, and human understanding is important since AI/ML increasingly affect us and our decisions~\cite{theCost,Webb2019tbn,o2016weapons,Vallor2016}. 

We here view contemporary ML as limited to local generalisation within a single task or well-defined set of tasks that only holds when the training data used is independent-and-identically-distributed (i.i.d). ML is then limited when this does not hold or when it comes to causal inference and out-of-distribution (o.o.d) generalisation~\cite{Chollet2019,scholkpf2021}. 

The human capability to generalise knowledge builds, as a contrast, on that we can formulate explanations using causal relations and generalise via concepts. Concepts are then building blocks for thoughts, blocks that are connected via relations forming explanations that in turn can bring understanding~\cite{sep-concepts, sep-understanding}.

Human understanding and trust in ML concerns not only \textit{understanding} promoted decisions\footnote{The output from an ML system in the form of classification, recommendation, prediction, proposed decision or action}, but also, evaluating these decisions in relation to limitations built into the ML model. Limitations are introduced in ML systems by humans during the design phase, for example; what to model, choice of algorithm, feature engineering, training data selection~\cite{Gillies2016}. The need for explanations to convey understanding is pronounced in more complex ML models~\cite{Lipton2016} and especially prominent in today's dominating technology: neural networks. 

Our approach towards understanding ML decisions builds on connecting human understandable concepts to the ML models knowledge representations with the goal of making them explicable. Below follows an outline of the perspectives on explanations used in this paper.

%\subsection{Explanations and objectivity} 
We use Hilton's~\citeyearpar{Hilton1990a} definition that explanations in a human context is a three-place predicate: \textit{Someone} explains \textit{something} to \textit{someone}. A definition that focuses on the explanation as a conversation between the explainer and the explainee.

Additionally, the need for an explanation in a human context is often triggered by an event that is abnormal or unexpected from a personal point of view~\cite{Hilton1986,hesslow1988problem,Hilton1996}. Research in Explainable Artificial Intelligence (XAI)~\cite{Barredo,Guidotti2018,Biran2017,Gilpin2019,Adadi2018,Hoffman2020}, on the other hand, are less concerned with \textit{who} gives the explanation, to \textit{whom} it is given or \textit{why} it is needed~\cite{Miller2019} and has, in comparison, a more objectivising decontextualised stance to explanations.

We then aim towards a situation where humans, with domain knowledge (henceforth domain experts), can act as explicators and formulate an explanation based on human understandable knowledge representations extracted from the neural network as concepts in line with Hiltons's~\citeyearpar{Hilton1990a} definition.

\begin{figure}[t]
\includegraphics[width=\linewidth]{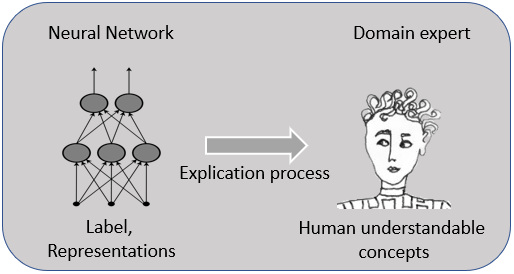}
\caption{Approach to explainability used in this work}
\label{fig:oversight}
\end{figure}

%%\subsection{Our contribution}
Figure~\ref{fig:oversight} is an overview covering the approach to explainability we aim for, which is a system where the neural network presents evidence for a decision in the form of knowledge representations in an explication process. The goal for the domain expert is to associate these representations to Human Understandable Concepts (HUC) and thus deepen their understanding of evidences for the decision and the models capabilities. The human can then be seen as the explainee in Hilton's definition, a person that aims to understand decisions, trust the model and use it to reach some goal.  

Focus in the work presented here is a targeted review that lifts out examples of existing literature on methods that aims to extract knowledge representations from neural networks. We are guided by the following research questions:
\begin{itemize}
    \item How do current methods extract internal knowledge representations in a neural network and map them to human-understandable concepts?
    \item Can Deductive-Nomological (D\nobreakdash-N) explanation taxonomy and causal types of explanations be useful in order to analyse what can be expected, and not expected, from knowledge representations in neural networks?
\end{itemize}

We answer the first question by organising the methods as global and local to discuss how HUC are induced by humans, either as knowledge priors added in the form of conceptual understanding or, during analysis of the explanans provided by the methods. We also find the D\nobreakdash-N taxonomy helpful, and that it, together with explanation types, opens up a generative path that makes it possible to better understand and discuss what we can expect from the type of machine learning we analyse.

The rest of the article is as follows: First, we present a more detailed description related to explanations and concepts, which is complemented by a description of the literature selection process followed by our review. The results are then deepened using our theoretical approach followed by a discussion related to our results. The work ends with a conclusion section. 

\section{Background and foundational concepts}
In this work we envision a situation with at least one domain expert that has the capability to understand and value knowledge representations extracted from a neural network. This prerequisite has the advantage of picturing a domain expert in the loop, a person that can be trained in scientific thinking and can relate to scientific explanations. The approach is also useful for persons not trained in scientific thinking since both mundane and scientific explanations aim at answering a \textit{why} or \textit{what if} question, with the difference that scientific explanations, in the D\nobreakdash-N case, aim towards objectivity and adds rigour to the answer. 

%\subsection{Concepts}
\label{sec:background concept}
According to \citet{Murphy2016} there are no exemplar theories of concepts. What can be generally agreed on is that concepts are named using a referent, for example, {\small\fontfamily{pcr}\selectfont GOLD}, and that a concept, like {\small\fontfamily{pcr}\selectfont GOLD}, can be grasped by believing that it has some properties as, a specific shine, value or its malleability. Peoples’ beliefs, related to concepts and their properties, can be both false and incomplete and, additionally, contain both causal and descriptive factors~\cite{Genone2012}. \citet{Ghorbani2019} deem that concepts, in relation to the domain modelled by a neural network, should be meaningful, coherent and important. 

This then implies that the domain expert's understanding of the trained model's behaviour is built on concepts that are derived from internal knowledge representations in a process that can be viewed as parallel to Carnap's \citeyearpar{carnap1962logical} theory of explications. The explication process is then the transformation or replacement of an imprecise concept (explicandum) with new concept(s) (explicatum/explicata). The new concepts adhere to the criteria of being similar to the imprecise concept but more exact, fruitful and simpler~\cite{justus2012}. These Human Understandable Concepts (HUC) can then be Disentangled (HUDC) if they are not confounded and they don't depend on spurious correlations.

For the work presented here, we imagine an explication process that refines and map the label, seen as a concept and internal knowledge representations, to human understandable concepts, with the goal of global understanding related to the model’s behaviour. These explicated concepts then aim to bridge the gap between the model's knowledge representations and the fraction of the real world it models. 

We use {\small\fontfamily{pcr}\selectfont kind}-related to refer to an abstract class of concepts, for example {\small\fontfamily{pcr}\selectfont SWAN, GOLD, DOG} or {\small\fontfamily{pcr}\selectfont DOTTED}. We use {\small\fontfamily{pcr}\selectfont entity} to refer to {\small\fontfamily{pcr}\selectfont kind} instances that are concrete particulars existing in time and space. For example, the {\small\fontfamily{pcr}\selectfont kind} concept {\small\fontfamily{pcr}\selectfont ZEBRA} can be explicated and refined by connecting it to the sub-concepts {\small\fontfamily{pcr}\selectfont HORSELIKE} and {\small\fontfamily{pcr}\selectfont STRIPED}. We then in this example, use two {\small\fontfamily{pcr}\selectfont kind}-related sub-kinds to create a causal explanation connected to the kind {\small\fontfamily{pcr}\selectfont ZEBRA}. We can then train a neural network using for example labelled images as {\small\fontfamily{pcr}\selectfont data}, so it can classify images containing {\small\fontfamily{pcr}\selectfont ZEBRAs} and {\small\fontfamily{pcr}\selectfont HORSEs} as HUDC. If there are a sufficient amount of data the network will generalise and be able to classify unseen images picturing {\small\fontfamily{pcr}\selectfont ZEBRAs} correctly. 

We can alternatively train a neural network using a core relation between {\small\fontfamily{pcr}\selectfont entities} to separate and classify, for example, individual {\small\fontfamily{pcr}\selectfont ZEBRAs}. The internal knowledge representations learned by the neural network will then relate to {\small\fontfamily{pcr}\selectfont ZEBRA} instances and, for example, be explicated as the HUDC {\small\fontfamily{pcr}\selectfont SCAR}, {\small\fontfamily{pcr}\selectfont BLURRED STRIPES} and {\small\fontfamily{pcr}\selectfont MARE}. We denote this type of concepts {\small\fontfamily{pcr}\selectfont entity}-related since they follow the instance.

%\subsection{Explanation types}
\label{sec:explanation types}
In this work  we, in line with Pearl~\citeyearpar{Pearl2019,Bareinboim2020}, define three types of explanations that answers to different types of \textit{what if}-questions: 
\begin{itemize}
    \item Association that answers to \textit{What if I see?}
    \item Interventional explanations answers to \textit{What if I do?}
    \item Counterfactual explanations that answers to \textit{What if I had done?}
\end{itemize}
At the first level we are only concerned with associations, e.g. regularities in the observation, and no understanding of cause and effect is needed. Interventional and counterfactual explanations builds on imagining alternative outcomes based on counterfactual causes introduced by consciously changing the prerequisites for the decision in question. This requires a causal model of the phenomena, a model that can be used to falsify a claim that make statistical sense. To use a classical example, a causal model that depicts why it is the sun that makes the roster to crow even if the crow preludes the sunrise.

The type of knowledge representations that can be created in a neural network are based on associations between {\small\fontfamily{pcr}\selectfont data} and a label, a label that then belongs to one of two D\nobreakdash-N categories: {\small\fontfamily{pcr}\selectfont entity} or {\small\fontfamily{pcr}\selectfont kind}. The trained ML model is then built using inductive statistical data and can consequently only answer to a \textit{What if I see?}-questions. This question is then answered by presenting the label and accuracy measurement. This can be sufficient in a static well-defined setting, but if the explainee wants a deepened understanding of how sensitive the decision is, for example, concerning the  {\small\fontfamily{pcr}\selectfont STRIPEDNESS} concept, we need to contrast the decision by using alternative input data in the form of counterfactual or semi-factual causes~\cite{Akula2020,kenny2021}. Intervention or \textit{What if I do?}-questions implies this type of doing and relies on causal understanding. On the top rung of Pearl's~\citeyearpar[p. 28]{pearl2018} ladder of causation are counterfactual explanations (\textit{What if I had done?}), that also builds on causality, but additionally also on a capability to imagine an alternative reality that would have manifested itself if another decision were taken in a given situation.

%\subsection{Explanation categories}
\label{sec:explanation categories}
In this work, we are interested in answering \textit{what if} questions, related to an ML decision and we use Deductive-Nomological (D\nobreakdash-N)~\cite{hempel} explanations to introduce rigour and structure to the answer. As outlined in Figure~\ref{fig:oversight}, we also leave it to the domain expert to account for or get insights into confounded features and causal relations. 

In line with Overton~\citeyearpar{Overton} we structure D\nobreakdash-N explanation using the following categories: {\small\fontfamily{pcr}\selectfont theory, model, kind, entity} and {\small\fontfamily{pcr}\selectfont data}. Neural networks build knowledge inductively using {\small\fontfamily{pcr}\selectfont data} and labels as a referent to concepts related to {\small\fontfamily{pcr}\selectfont entities} and/or {\small\fontfamily{pcr}\selectfont kinds}. The model trained in this process is then not a {\small\fontfamily{pcr}\selectfont model} in a D\nobreakdash-N sense, since a D\nobreakdash-N {\small\fontfamily{pcr}\selectfont model} is justified using a {\small\fontfamily{pcr}\selectfont theory} that articulate relations between {\small\fontfamily{pcr}\selectfont kinds}.  

\begin{figure}[t]
\includegraphics[width=\linewidth]{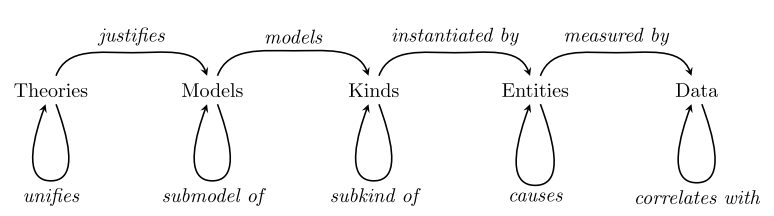}
\caption{Structure of D-N-explanation, used by permission from Overton.}
\label{fig:overton}
\end{figure}

Overton~\citeyearpar{Overton} outlined a generalised structure for scientific explanations (See Figure~\ref{fig:overton}). Since we use neural networks only the categories {\small\fontfamily{pcr}\selectfont kind}, {\small\fontfamily{pcr}\selectfont entity} and {\small\fontfamily{pcr}\selectfont data} are involved and can be used to build explanations. The deductive part of the explanation is missing and this delimits the explanadum. Instead of a {\small\fontfamily{pcr}\selectfont theory} justifying a {\small\fontfamily{pcr}\selectfont model} the ML model is built using statistical {\small\fontfamily{pcr}\selectfont data} and we equal this to a law in a D\nobreakdash-N sense. The structure of scientific explanation that builds on a core {\small\fontfamily{pcr}\selectfont kind\nobreakdash-data} and {\small\fontfamily{pcr}\selectfont model\nobreakdash-data} relation can be seen in Figure~\ref{fig:kind-data}. In those figures the explanadum is denoted quality B and the explanans, that are the evidences for a decision are denoted quality A. A complete D\nobreakdash-N explanation ({\small\fontfamily{pcr}\selectfont theory\nobreakdash-data}) can be seen in Figure~\ref{fig:theory-data}.      

%\subsection {Extracting concepts}
A majority of the research reviewed in this work uses raw input data in the form of images and we will not to any large degree discuss input data in other forms. In relation to the issue of classification of explanation methods we adhere to \cite{Gilpin2019} that organise the methods in three categories: 
\begin{itemize}
    \item Methods based on the processing of data, which we denote Feature Based Attribution (FBA).
    \item Methods based on Internal Knowledge Representation (IKR).
    \item Methods that aims to automatically create understandable explanations.
\end{itemize}
We will focus our work on the first two categories and leave it to humans with domain knowledge to create explanations based on FBA and/or IKR explanans. Related to the discussion in the introduction we find it difficult to imagine automatically created explanations valid in a human context in general, without any restrictions related to the domain, the context or, as in our approach presume human explicators with domain knowledge (Se Figure~\ref{fig:oversight}).  

%\subsubsection{FBA}
Well-cited FBA methods are for example Grad-CAM~\cite{Selvaraju2016}, LIME~\cite{Ribeiro2016} and SHAP~\cite{Lundberg2016}. These methods are local in the sense that they are used to reveal evidence for a decision for a specific input (f.x. an image). LIME and SHAP rely on the creation of a local interpretable substitute model (for example a linear model) using perturbation on input features to infer which features the classification are sensitive for. For images, this approach can be used to grey out areas and expose \textquote{super-pixels}, e.g. areas in the picture that the classification is sensitive for. Grad-CAM belongs to a category of methods that uses gradients to attribute model output to layers in the model or to input features. Grad-CAM specifically uses the last convolutional layer and therefore combines high-level semantic information with spatial information. The methods being local does not rule out that they over time, with usage in different situations, can result in a global understanding of the model and trust in its decisions. The analogy here being, for example, trusting a dog interacting with your kids, a trust that are built over time using singular specific situations. Related to concepts, these methods exposes the relation between the internal knowledge representations and a specific image. This implies that for an image that belongs to the assumed i.i.d training data the methods can reveal information on learned representations.

%\subsubsection{IKR}
Methods that builds on extracting IKR are global since they reflect the neural networks overall learning process. A network is forced, during training, to learn disentangled representations in each layer in the form of vectors connected via weight matrices, these vectors are then generalisations on some representation level~\cite{benigio2021}. It is these generalisations that potentially can map to concepts and they tend to get more complex with the depth of the layers. Therefore more basic concepts like colours, patterns and shapes are represented in the early layers and concepts like {\small\fontfamily{pcr}\selectfont GENDER} and {\small\fontfamily{pcr}\selectfont HORSE} in later. An influential method, Concept Activation Vectors CAV~\cite{Kim2018}, for extracting knowledge representations, or variants of it, is used in a majority of the reviewed papers. CAV can be used for example to expose images from a training set sensitive to the concept {\small\fontfamily{pcr}\selectfont STRIPED} or to reveal the correlation between vectors that represent simpler concepts and more complex concepts, for example, between the colour {\small\fontfamily{pcr}\selectfont RED} and {\small\fontfamily{pcr}\selectfont FIRE TRUCK}.

%%\subsection {Methodology}
\label{sec:methodology}For this review we mainly use \citet{webster2002} and \citet{knopf2006} as guidelines. For our review we were interested in a representative selection of relevant research, useful to indicate the applicability of our theoretical approach. The XAI field is a vast area and we experimented with search criterion that were general enough to result in a, for our purpose, useful and manageable selection of research articles.  We searched in the abstract, title or keywords in research published between 2018 and September 2021 that has the term \textquote{understandable} within three words before the term \textquote{concept} and that the abstract, title or keywords contained either \textquote{neural network} or \textquote{deep learning}. By limiting to the search like this we could target papers that had the intended focus and still use more recent XAI methods. The challenge with the search is otherwise that our search terms are to general, especially the concept \textit{concept} that is used in many ML related fields. We searched IEEE Xplore, Scopus, Web of Science and ScienceDirect and got 13 relevant hits. From these we removed work related to mathematically understandable and not human understandable. 

In the end, we had nine papers targeting the area. We analysed the papers given our approach and a setting where a human domain expert explicates the internal knowledge representations exposed to create explanations. We then applied our theoretical lens centred on concepts, the D\nobreakdash-N-model and the types of explanations mentioned above. At the review phase, we organised the research concerning the type of explanation the work presented aimed it for.  

\section{Review}
In this section, we present our review results that build on our targeted search question, theoretical lens and the methodological approach presented earlier. 

\begin{table*}[t]
        \begin{center}
        \footnotesize
        %\tiny
        \begin{tabular}[b]{|| p{8em} | p{4em}| p{16.5em} | p{16.5em} | p{4em} ||} 
        \hline
         &\multicolumn{4}{c|}{{\fontfamily{pcr}\selectfont Core relation} } \\
         \hline
         \textbf{Explanation type}
         & 
         \centering{\fontfamily{pcr}\selectfont theory-\newline data}
         & 
           \centering{\fontfamily{pcr}\selectfont model-data}
         & 
          \centering{\fontfamily{pcr}\selectfont kind-data}
         &
            {\fontfamily{pcr}\selectfont entity-\newline data}  \\
         \hline\hline
            \textbf{Counterfactual} \newline What if I had done?
            & 
             &
             &
              *Expose temporal factors between correlated concepts~\cite{Mincu2021}.
             &
             \\
            \hline
            \textbf{Intervention} \newline What if I do? 
            &
            &
            *Expose decision tree that containing main concepts used for the decision~\cite{Elshawi2021}. \newline 
            *Expose first order logic for a decision~\cite{Rabold2020}.
            & 
            *Expose typical vs -atypical images~\cite{Chen2020,Lucieri2020}. \newline
            *Relate decision to predefined disentangled concepts~\cite{Chen2020}.
            &
            \\ 
            \hline
            \textbf{Association} \newline What if I see?
            & 
            &
            *Expose human machine learning epistemic alignment \cite{Natekar2020,Lucieri2020,Yeche2020}. \newline 
            *Expose necessary and sufficient concept attribution~\cite{Wang2020}.
            &
            *Expose concept-typical images ~\cite{Lucieri2020,Ghorbani2019}. \newline
            &
            \\
        \hline\hline
\end{tabular}
\end{center}
\caption{Categorisation and structure of explanations the reviewed articles \textbf{aim} for. %It should be noted that the intervention and counterfactual explanations builds on knowledge priors added to the model.
}
\label{tab:the stuff}
\end{table*}

%\subsection{Two central approaches to concept mining}
As the search criterion is formulated the focus is on research that aims to connect internal knowledge representations in neural networks to HUC. Two approaches dominate in the reviewed papers, either unveiling HUC using FBA and local understanding, used in two papers~\cite{Wang2020,Natekar2020}, or more directly by using IKR~\cite{Yeche2020,Ghorbani2019,Elshawi2021,Mincu2021,Chen2020,Rabold2020,Lucieri2020}, used in the remaining seven papers. In both cases, the goal is to better understand the model from a global perspective via concepts. The two approaches are in most papers presented as a dichotomy between global and local understanding which, of course, is relevant if the system is not used over some time, or that a user of the system can experiment with a combination of explainability methods. In ~\citet{Ghorbani2019} the approaches are to some extent combined in that an initial segmentation of the image is performed and then the image segments are clustered using IKR to calculate the segment's importance in relation to the concepts they represent. This points towards interesting opportunities in combining methods to infer the best explanation where the explanandum reachable is represented by IKR and the explanans, as evidence for a decision, by FBA. 

Related to popular explainability methods: six of the nine papers mention CAV~\cite{Kim2018}, five mentions Grad-CAM~\cite{Selvaraju2016}, one LIME~\cite{Ribeiro2016} and not any mentions SHAP~\cite{Lundberg2016}. 

%\subsection{Epistemic alignment}
The research reviewed in the medical field~\cite{Natekar2020,Lucieri2020,Yeche2020} sticks out compared to research reviewed in the non-medical field. One central goal in the medical domain is the search for alignment between human concepts and IKR to create trust in the model decisions. In \citet{Natekar2020} an alignment between the human concept identification process and the same process in a neural network is highlighted and in \citet{Lucieri2020} an alignment between disentangled concepts identified by the neural network and concepts routinely used by dermatologists is unveiled. In \citet{Yeche2020} the consistency of a concept over the layers in the network is exposed. 

Explanans useful to underpin a decision in this domain has their base in artificially created 2D images and we can hypothesise that there is a substantial overlap between the explanandum reachable for the ML system and the explanandum reachable for the human.  
 
%\subsection{Epistemic misalignment}
The work by \citet{Ghorbani2019} briefly discusses how machine identified HUC can be misleading and include concepts not aligned with human understanding. For example, that the player's jerseys in basketball is a more important concept than the ball for predicting the sport in question. The work by \citet{Wang2020} lifts similar concerns related to that concepts deemed by the ML system to be \textit{sufficient} and/or \textit{necessary} can be misleading if, for example, images used relates to complex situations or situations not reflected in the training data. The example made in that paper is: If training data for traffic signs only contains stop signs on poles, the pole can be deemed as \textit{necessary}. Consequently, a stop sign without a pole can be classified as a false negative with potentially serious implications. These situations can be viewed as a lack of overlap between the explanandum reachable for the human vis-a-vis the explanandum reachable for the ML system.

%\subsection{Explanation types}
In our work, we discuss explanations that build on association, intervention and on counterfactuals. In the work we reviewed, explanation types are not discussed in-depth, instead, they are treated more implicitly as a part of self-evident background knowledge. In our work we are interested in a deepened discussion on the role of explanations in ML to better understand if and how explanation types can be a useful and actionable tool to understand an ML model's abilities and limitations. An initial step is to arrange the reviewed research in rows, reflecting the different types of \textit{what if}-questions the research targets (See Table~\ref{tab:the stuff}). Below follows a categorisation of the explanation types, starting with association being least complex and placed at the bottom row in Table~\ref{tab:the stuff} thus organising the table in line with Pearl's~\citeyearpar{Pearl2019} causal hierarchy. Some of the reviewed articles appear in more than one place since they present the use of XAI methods with different goals in parallel experiments.   

\subsubsection{Association (What if I see?)} 
One example of direct association, is research that exposes images similar to an input image for human comparison to a skin lesion concept using IKR~\cite{Lucieri2020}. \citet{Ghorbani2019} segments images and then uses IKR to align concepts for human comparison. In \citet{Wang2020} an algorithm is created to calculate if a HUC is sufficient and/or necessary for a decision. Three research papers find alignment between how humans and machines learn as their central explanans~\cite{Natekar2020,Lucieri2020,Yeche2020}. \citet{Natekar2020} sticks out in that the explanans extracted aims at creating trust in the ML decision process by showing that the process is aligned with a human decision process. We arrange these research papers as an association since they do not offer any alternative decision and instead focus on presenting evidence for a human that can be used to increase trust in the system or a decision.     

\subsubsection{Interventions (What if I do?)}
In the reviewed research this type of question is addressed by: exposing typical and atypical images related to a concept~\cite{Chen2020,Lucieri2020}, relate a decision to one of a number of predefined disentangled concepts~\cite{Chen2020} or building simplified models over the decision logic~\cite{Elshawi2021,Rabold2020}. We arrange these systems as aiming for intervention in relation to their input data and their labels since they present both contrastive and non-contrastive explanans thus making it possible for a domain expert to construct \textit{what if} explanations. Knowledge priors added are, for example, the selection of disentangled concepts or using a decision tree as the structure for a contrastive explanation.   

\subsubsection{Counterfactuals (What if I had done?)} 
Since counterfactual explanations build on a capability to imagine alternative futures, from an historical point in time, there is a need for temporal data for this type of explanation. In the reviewed work there is one example based on electronic health records and a comparison between how a specific treatment at a certain situation, for example, antibiotics treatment, can be evaluated in comparison to alternative treatment~\cite{Mincu2021}.

\subsubsection{Explanation categories}
If we view the reviewed research through the lens of D\nobreakdash-N explanations the explanations aims for either a {\small\fontfamily{pcr}\selectfont model-data} or {\small\fontfamily{pcr}\selectfont kind-data} relation. {\small\fontfamily{pcr}\selectfont Data} are in all cases images except in \citet{Mincu2021} where temporal tabular data from electronic health records is used. Below we analyse the work reviewed in relation to their core relation.  

\begin{figure}[t]
    \includegraphics[width=\linewidth]{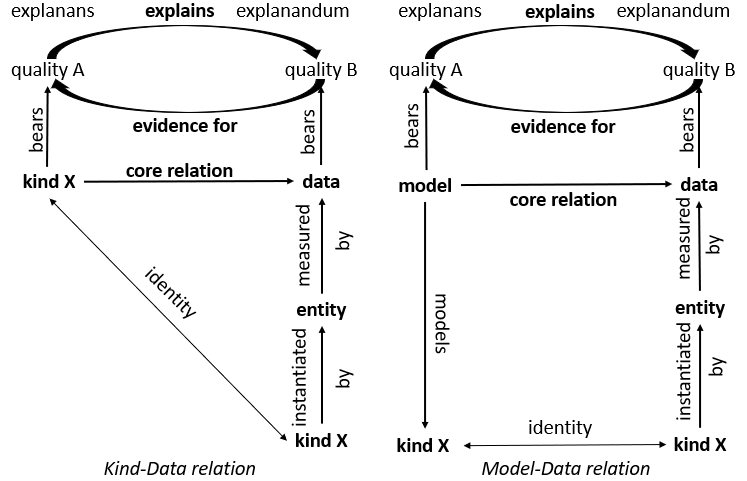}
    \caption{The two explanation structures the review articles aim for.}
    \label{fig:kind-data}
    \end{figure}

%\subsubsection{Core relation {\small\fontfamily{pcr}\selectfont kind-data}}
In Figure~\ref{fig:kind-data}, to the left, the structure of a {\small\fontfamily{pcr}\selectfont kind-data} explanation is presented. As discussed earlier we view the trained model as a law under scrutiny and not a {\small\fontfamily{pcr}\selectfont model} in D\nobreakdash-N sense. For example, in \citet{Lucieri2020}, that focuses on identifying skin lesions, the images presented as similar to the image to be explained, together, with the trained model create the explanans that are a selected part of the available explanandum. This explanandum then answers a question of the form: This instance belongs to the concept \textbf{quality a} (a specific skin lesion concept) presenting these similar images as evidence/explanans for the decision (\textbf{quality b}). If a human with domain knowledge finds that these evidence sufficient, perhaps by combining them with other factors as experience, known sub-{\small\fontfamily{pcr}\selectfont kind} concepts or data not included in the training data then an explanation that builds on causality (if C then D) valid in the real world can be formulated by the domain expert.  

In  one example by \citet{Chen2020} the correlation between five disentangled {\small\fontfamily{pcr}\selectfont kind} concepts: {\small\fontfamily{pcr}\selectfont BEDROOM, AIRFIELD, STABLE, BOAT DECK} and {\small\fontfamily{pcr}\selectfont INDOOR LIBRARY} and seven disentangled sub-{\small\fontfamily{pcr}\selectfont kind} concepts: {\small\fontfamily{pcr}\selectfont AEROPLANE, BED, BENCH, BOAT, BOOK, HORSE, PERSON} is part of a proof of concept experiment. Here the explanadum consists of the classified {\small\fontfamily{pcr}\selectfont kind}, the trained model and sub-{\small\fontfamily{pcr}\selectfont kinds} pictured and not pictured in the image. A contrastive explanation can then be formulated around the inner workings of the model for example that: since the sub-{\small\fontfamily{pcr}\selectfont kinds} identified are {\small\fontfamily{pcr}\selectfont PERSON} and {\small\fontfamily{pcr}\selectfont BED} the model classifies the picture as a {\small\fontfamily{pcr}\selectfont BEDROOM} and not a {\small\fontfamily{pcr}\selectfont STABLE}. This explanation is contrastive from the trained model's perspective in the sense that it explains the classification and that it is likely that images containing a bed and a person will be classified as bedrooms. Here, again, knowledge priors in the form of selection of {\small\fontfamily{pcr}\selectfont kinds} and sub-{\small\fontfamily{pcr}\selectfont kinds}, but also training data and model selection, together delimits the explanadum available. So even if the form of the explanation can be classified as intervention the explanation is not causal in the sense that it holds in a real world context, instead it gives insights in how the trained model associate input data with outputs. 
The same holds for the proof of concept in \citet{Mincu2021} where a system in the hands of a person holding medical expertise can be a tool useful to create counterfactual explanations. The explanans presented by the system, together with other explanantia, can open up to better understand the consequences of an alternative historical decision. For example that it is probable that using a different type of antibiotics on women than men will make women recover faster.

%\subsubsection{Core relation {\small\fontfamily{pcr}\selectfont model-data} }
In Figure~\ref{fig:kind-data}, to the right, the structure of a {\small\fontfamily{pcr}\selectfont model-data} explanation is presented. We placed the work that compares the human decision process with the ML decision process as a core relation between {\small\fontfamily{pcr}\selectfont model} and {\small\fontfamily{pcr}\selectfont data} since the aim is to use an overlap as an explanan and useful {\small\fontfamily{pcr}\selectfont model} over a trustworthy decision process~\cite{Natekar2020,Lucieri2020,Yeche2020}. In \citet{Wang2020} calculations of sufficient and necessary causes are used to explain a decision. Knowledge priors added here are then under which conditions a presupposed value of an alignment to human learning is useful as an explanan and under which conditions it is possible to calculate sufficient and necessary cause in an inductive learning process. 

In two reviewed work a {\small\fontfamily{pcr}\selectfont model}, in a D\nobreakdash-N sense, over the decision process is created. In the work by \citet{Rabold2020} sub-{\small\fontfamily{pcr}\selectfont kinds} are automatically identified and used to build first order logical rules covering spatial relations in images (the relative placement of {\small\fontfamily{pcr}\selectfont EYES, MOUTH} and {\small\fontfamily{pcr}\selectfont NOSE} useful to identify a {\small\fontfamily{pcr}\selectfont FACE}). In \citet{Elshawi2021} a decision tree based on sub-{\small\fontfamily{pcr}\selectfont kinds} is constructed and used to explain classification of pictures, answering contrastive questions like: Why is this image classified as a {\small\fontfamily{pcr}\selectfont COAST} and not a {\small\fontfamily{pcr}\selectfont MOUNTAIN}?. 

The work reviewed does not include any research that uses a {\small\fontfamily{pcr}\selectfont theory-data} relation or a {\small\fontfamily{pcr}\selectfont entity-data} relation (See Table \ref{tab:the stuff}). Related to {\small\fontfamily{pcr}\selectfont theory-data} it can be argued that using a surrogate {\small\fontfamily{pcr}\selectfont model} implicitly presumes that the proposed decision can be explained using, in this case, a decision tree or first order logic~\cite{Rabold2020,Elshawi2021}. 

There is in the reviewed work no trained model that focus on {\small\fontfamily{pcr}\selectfont entity-data} relations, relations that a neural network can learn well in a similar fashion as it can learn {\small\fontfamily{pcr}\selectfont kind-data} relations. {\small\fontfamily{pcr}\selectfont Entity}-related concepts follows the instance and can be related to ageing or wear and tear, for example, {\small\fontfamily{pcr}\selectfont SCRATCHES, SPLINTERS} or {\small\fontfamily{pcr}\selectfont MARKINGS} and can be useful to, for example identify objects and estimate ageing~\cite{Holmberg2021SMC}.

%\subsection{User test}
In line with \citet{Tjoa2019} we find a lack of user studies in the work we reviewed. The studies conducted are limited even if they address a mundane domain where users are readily available~\cite{Elshawi2021,Ghorbani2019}. The lack of studies in explainable AI is notable since the research heavily relies on human traits and abilities to, for example, compare images for similarity and infer disentangled sub-{\small\fontfamily{pcr}\selectfont kind} concepts. Concrete examples from the work we reviewed includes, look at images picturing beds from different environments and relate them to the concept {\small\fontfamily{pcr}\selectfont BED}~\cite{Chen2020} or infer that the tail is sufficient evidence to classify a {\small\fontfamily{pcr}\selectfont KOMODO LIZARD} but that the rocks it rests on and the body are necessary evidence~\cite{Wang2020}. 

The reviewed work exposes a wealth of undefined expectations on what the explainee can infer from the evidence for a decision exposed by the explainability methods. Specifically, the explainee is expected to understand the model's limitations and be aware that explanations produced reflect the training data, the architecture used and that it only is an incomplete overlay on the reality it models. The possibility for an explainee to fathom this difference is crucial if we are not only interested in the more introspective project of evaluating the model built as such but interested in applying proposed decisions in the real world. For example, to evaluate if a predefined named concept selection is relevant in relation to the domain targeted and the decision promoted or, alternatively, if the decision is due to exposure to o.o.d and non i.i.d data. For example, evaluate if the predefined concepts chosen: {\small\fontfamily{pcr}\selectfont BED}, {\small\fontfamily{pcr}\selectfont SINK}, {\small\fontfamily{pcr}\selectfont SEA}, {\small\fontfamily{pcr}\selectfont TREE}, {\small\fontfamily{pcr}\selectfont HIGHWAY} are the best one to evaluate the classification as a image picturing a {\small\fontfamily{pcr}\selectfont COAST}~\cite{Elshawi2021}.

\section{Discussion}
In this section, we discuss the review results using our theoretical lens. Initially, the centrality of concepts is lifted followed by a comparison with the systems we aim for using scientific instruments as a comparison and base for the discussion. We then focus on the central notion of causality and the role it has related to explanations, this is followed by a discussion on training data distribution. The section ends with limitations and a summary section.

%\subsection{XAI method reflection}
It is perhaps not surprising that methods that build on extracting IKR dominate the review papers since they aim for global understanding more \textquote{by design}. FBA methods are local in that they compare one decision with the trained ML models internal knowledge representations. In our setup, FBA methods can be used to falsify the model and get a deeper understanding related to outliers in non-i.i.d data and unknown domain related concepts in o.o.d data. IKR methods, on the other hand, give insights that are more general and \textquote{typical} for the trained model. This point towards the need for user-studies that combines IKR and FBA methods in studies that aim towards understanding the model from a global meta-perspective similar to how we \textquote{understand} and trust companion species. It is somewhat surprising that in other XAI papers well-cited research, LIME and SHAP, are relatively invisible in the selected articles. One reason can be that they, in creating local surrogate models (f.x. linear), adds an extra layer that needs to be interpreted to get global understanding. In our setting these surrogate models can be faster to interpret since they simplify and can allow for the domain expert to experiment and search for decision boundaries or get a general overview of the knowledge representations learned by the ML system.    

%\subsection{Concept} 
Explicating and identifying concepts are in all cases in the reviewed research done using human knowledge. The incomplete world model in the trained models becomes apparent in, for example, \citet{Ghorbani2019} where images of ocean water that is {\small\fontfamily{pcr}\selectfont CALM}, {\small\fontfamily{pcr}\selectfont WAVY} and {\small\fontfamily{pcr}\selectfont SHINY} are identified as separate concepts and not as sub-concepts of {\small\fontfamily{pcr}\selectfont OCEAN}. Developing concept ontologies, concept formulation and reformulation are then seminal and support the position we take here, that a human with domain expertise interacting with the system needs to be part of any system, based on inductive learning, used in a non-stationary context. 

%\subsection{Scientific instrument} 
In the radiology related research \citet{Yeche2020} and \citet{Natekar2020} the approach can be seen as an incremental development of scientific instruments in line with previous technological progress within a well-defined usage domain in the hands of domain
experts~\cite{Roscher2020,Karpatne2017}. The research focuses on shared hermeneutic 2D images that have a substantial explanandum overlap between the ML system and the domain expert. The work reviewed that targets the medical domain focus on actionable explanations related to the decision and less on explaining the ML models inner workings. This is an indication that the current focus in ML on large static data sets needs to be complemented with datasets that has more overlap with human understanding of how the world is constituted, its diversity and contextual dependence. Increased focus on {\small\fontfamily{pcr}\selectfont entity-data} relations can then complement the current objectivising {\small\fontfamily{pcr}\selectfont kind-data} focus, and open up an interesting path towards more contextual and small-scale usage of ML systems, systems that then can add value to humans in context.

%\subsection{Causal relations}
In this work, we pay special attention to what ML, in the form of neural networks, \textit{cannot} do based on its statistical inductive learning approach. The epistemic consequences are originally formulated by David Hume as the problem of induction~\cite{sep-induction-problem}, popularised as the black swan problem, a problem that cannot be solved using more data. Since ML/AI of today is void of understanding, and only handles local domain generalisation~\cite{Chollet2019}, we have to be mindful of the black swans these systems do not see and can hide for us even if we as humans, and understanding seeking animals, are aware of them~\cite{Prasada2014}. To combine these information processing artefacts that ML systems are, with humans seeking understanding, we need systems that can explain themselves or, as the focus in the work presented here, find protocols so humans can understand and compensate for ML systems shortcomings. 

The reviewed research that aims for interventional or counterfactual explanations in Table~\ref{tab:the stuff} rely on causal relations. These contrastive and counterfactual explanations are then only valid for the ML model in isolation. We find that there, in the reviewed work, is a lack of discussion related to this, if the goal is to create explanations applicable in the domain the ML system targets. For example calculation of necessary and sufficient causes~\cite{Wang2020} has to be evaluated towards how well the training data reflects the context it will be used in, and if the training data carries this subjective information in a form that is not only statistical. 

The approach promoted here that views the ML system as a tool in the hands of a domain expert is one path forward underpinned by limitations unveiled in the reviewed work. Especially the {\small\fontfamily{pcr}\selectfont theory} and {\small\fontfamily{pcr}\selectfont model} categories need to be part of the discussion so deductive reasoning can be included and implemented in the ML-system or by using human capabilities. This would then be an approach that makes it possible to understand and challenge a, by the ML system, promoted decision.

%\subsection{Limitations and clarifications}
Our choice to use the D\nobreakdash-N model and Overton's~\citeyearpar{Overton} schematic overview delimit the type of explanations we aim for and exclude pragmatic, inductive statistical explanations, teleological and historical. Also, we have not deepened the discussion around the difference between description, justification, argument and explanation~\cite{Woodward2019,salmon1989}. 
As for concepts, we see them as central building blocks for thoughts and we are aware of that our understanding of the concept {\small\fontfamily{pcr}\selectfont CONCEPT} is not here well-grounded from an ontological and epistemological perspective.

%\subsection{Future research}
\begin{wrapfigure}{r}{0.5\linewidth}
    \begin{center}
        \includegraphics[width=\linewidth]{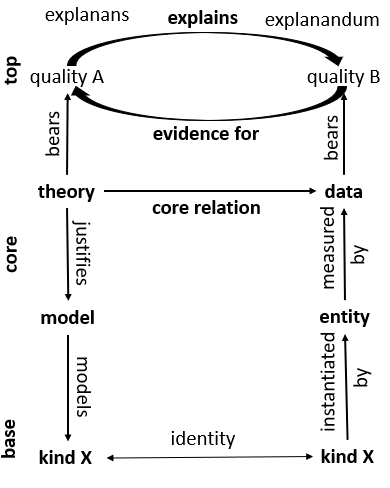}
        \caption{{\small\fontfamily{pcr}\selectfont theory-data} relation based on {\small\fontfamily{pcr}\selectfont kind} identity.}
        \label{fig:theory-data}
    \end{center}
\end{wrapfigure}

Our framing using scientific explanations and types of explanations illuminate an ambiguity in the reviewed work concerning the goal of the explanation. Is the goal understanding the ML model or understanding the domain modelled by the ML system? We can also see that this ambiguity is less pronounced when there is a large explanandum overlap between the reality the ML model models and the reality we as humans perceive. An interesting path forward is then to use scientific explanations, for example, using a complete general structure for D\nobreakdash-N explanations~\cite[pg. 17]{Overton} (See Figure~\ref{fig:theory-data}). Using this structure to formulate theories and falsifiable hypothesis similar to a traditional research process is one path towards evaluating how well a trained ML model models the usage domain. By doing this we move away from the idea that more data solves the problem of induction and instead treat ML systems as tools that can mediate better understanding. By complementing ML systems that builds on large data sets with {\small\fontfamily{pcr}\selectfont theories} that can be translated to {\small\fontfamily{pcr}\selectfont models}, in the form of, for example, causal graphs, algorithms and logic, we can add a needed {\small\fontfamily{pcr}\selectfont model} layer to these systems. Addressing these explanation structures is an important future focus that can create systems that can be challenged and possible to learn from.          

%\subsection{Summary}
In this work, we take an outside perspective in relation to a trained ML system and we find similarities with a scientific process that uses hypothesis and theories that are possible to challenge, improve and refine in relation to the domain targeted. Additionally we lift out a number of areas, the importance of concepts, causality, data shift and explanation types and explanation categories, essential to make these systems falsifiable.    

\section{Conclusion}
ML systems increasingly affect many aspects of human life, gaining trust in their decisions is a central and active research area. \textit{What if}-questions and the centrality of concepts is the focus for this review where we examine how concepts are extracted from a neural network. We presuppose a situation where a human, with domain knowledge, use concepts to answer why-questions. In our review, we use the structure of D\nobreakdash-N explanations and three types why-questions, \textit{What if I see?}, \textit{What if I do?} and \textit{What if I had done?} as an analytic lens to deepen and detail what we can expect, and not expect, from the research reviewed.

This review raises important questions on \textit{What is the goal for the explanation?} and \textit{What type of knowledge can be extracted from a neural network?}. Related to the first question we see the importance of differentiating between explanations that focus on a better understanding of the trained model, void of context, and those that focus on actionable decisions, useful in the deployment context. Related to the second question we see that the reviewed work, in many cases, aims for explanations that build on causal relations, that are required for these types of explanations, without discussing how these are added to the system.

We believe that studies that actively involve users can emphasise contextual dependence and refocus research in the area more towards the limitations of ML systems and consequently open up for an awareness of the societal and environmental impact these systems have when they are deployed.  
\bibliography{references} 
\end{document}